\documentclass[sn-basic,Numbered]{sn-jnl}

\newinsert\footinsA
\providecommand{\SetFootnoteHook}[1]{}
\makeatletter
\@ifundefined{DeclareNewFootnote}{\def\DeclareNewFootnote#1[#2]{}}{}
\providecommand{\headerps@out}[1]{}
\makeatother

\usepackage{graphicx}
\usepackage{booktabs}
\usepackage{float}
\usepackage{longtable}
\usepackage{tabularx}
\usepackage{array}
\usepackage{xcolor}
\usepackage{amsmath,amssymb,amsfonts}
\usepackage{amsthm}
\usepackage{fontspec}
\usepackage{ragged2e}
\usepackage{tikz}
\usepackage[section]{placeins}

\graphicspath{{./}}
\usetikzlibrary{arrows.meta,positioning}

\defaultfontfeatures{Ligatures=TeX}
\newcommand{\family}[1]{\textit{#1}}
\newfontface\persianfont[
  Path=./,
  Extension=.ttf,
  UprightFont=*,
  Script=Arabic,
  Scale=MatchLowercase
]{Vazir}
\newcommand{\persian}[1]{{\persianfont #1}}
\newcommand{\symfirst}[3]{\persian{#1} \family{#2} (#3)}
\newcommand{\symgloss}[2]{\family{#1} (#2)}
\newcommand{\symentry}[3]{\persian{#1} / \symgloss{#2}{#3}}
\providecommand{\doi}[1]{doi:\ #1}
\renewcommand{\arraystretch}{1.12}
\makeatletter
\let\ps@plain\ps@headings
\makeatother
\definecolor{workflowsea}{HTML}{4A89A8}
\definecolor{workflowteal}{HTML}{2F6F6D}
\definecolor{workflowgold}{HTML}{C58F2F}
\definecolor{workflowolive}{HTML}{6C8740}
\definecolor{workflowrust}{HTML}{B85D3D}
\definecolor{workflowplum}{HTML}{6F5D83}
\definecolor{workflowink}{HTML}{22313D}
\definecolor{workflowgray}{HTML}{5D6770}
\definecolor{workflowborder}{HTML}{DBE1E4}
\definecolor{workflowarrow}{HTML}{8F99A3}
\makeatletter
\newcommand{\manualfigcaption}[1]{%
  \vskip\abovecaptionskip
  {\figurecaptionfont #1\par}%
  \vskip\belowcaptionskip
}
\makeatother

\title[A Dynamic Atlas of Persian Poetic Symbolism: Families, Fields, and the Historical Rewiring of Meaning]{A Dynamic Atlas of Persian Poetic Symbolism: Families, Fields, and the Historical Rewiring of Meaning}

\author[1]{\fnm{Kourosh} \sur{Shahnazari}}
\author[1]{\fnm{Seyed Moein} \sur{Ayyoubzadeh}}
\author[2]{\fnm{Mohammadali} \sur{Keshtparvar}}

\affil[1]{\orgname{Sharif University of Technology}, \orgaddress{\city{Tehran}, \country{Iran}}}
\affil[2]{\orgname{Amirkabir University of Technology}, \orgaddress{\city{Tehran}, \country{Iran}}}

\abstract{Persian poetry is often remembered through recurrent symbols before it is remembered through plot. Wine vessels, gardens, flames, sacred titles, bodily beauty, and courtly names return across centuries, yet computational work still tends to flatten this material into isolated words or broad document semantics. That misses a practical unit of organization in Persian poetics: related forms travel as families and gain force through recurring relations.

Using a corpus of 129{,}451 poems, we consolidate recurrent forms into traceable families, separate imagistic material from sacred and courtly reference, and map their relations in a multi-layer graph. The symbolic core is relatively sparse, the referential component much denser, and the attachment zone between them selective rather than diffuse. Across 11 Hijri-century bins, some families remain widely distributed, especially \symfirst{شب}{Shab}{Night}, \symfirst{روز}{Ruz}{Day}, and \symfirst{خاک}{Khaak}{Earth}. Wine vessels, garden space, flame, and lyric sound strengthen later, while prestige-coded and heroic-courtly vocabulary is weighted earlier.

Century-specific graphs show change in arrangement as well as membership. Modularity rises, cross-scope linkage declines, courtly bridges weaken, and sacred bridges strengthen. Hub positions shift too: \symfirst{خرقه}{Kherqe}{Sufi Robe} gains late prominence, \symfirst{فرخنده}{Farkhondeh}{Blessed} and \symfirst{بنفشه}{Banafsheh}{Violet} recede, and \symfirst{ساغر}{Saaghar}{Wine Cup} stays central across the chronology. In this corpus, Persian symbolism appears less as a fixed repertory than as a long-lived system whose internal weights and connections change over time.}


\begin{document}

\maketitle

\section{Introduction}

Persian poetry has long been read as a culture of recurrent images. A rose is rarely only a rose, a cup rarely only a vessel, and the night sky rarely a neutral backdrop. In lyric, mystical, and courtly registers alike, familiar symbolic materials return in changing constellations. Their force depends on relation: what they cluster with, what they attract, and what historical pressures shape their reuse \cite{schimmel1992,seyedgohrab2011}.

That problem sits awkwardly within the usual computational scales of literary study. Token-based methods are useful for lexical profiling, stylistic comparison, and retrieval, but they split apart symbolic units that readers often treat together. Document-level semantics can recover broader thematic structure, yet they blur the distinction between imagistic symbols, sacred names, courtly titles, and generic semantic association. For Persian poetry, the useful scale lies between those poles: not the isolated form, and not the whole poem, but a family of related symbolic forms whose literary work becomes legible in aggregate \cite{moretti2013,jockers2013,underwood2019}.

We model that middle scale through symbolic families. A family is a traceable grouping of closely related surface forms supported by contextual and relational evidence. Working at this level reduces token fragmentation without losing lexical accountability. It also keeps sacred and courtly reference visible without letting it determine the same topology as imagistic symbolism.

The study asks three connected questions. Does family induction produce a cleaner symbolic inventory than token-only modeling? Does the system separate into imagistic, referential, and attachment structures? And across the mapped centuries, which families persist, which shift, and which change structural role inside the graph? Those questions matter both for Persian literary history and for computational work that needs a stable unit between word and document.

\section{Related Work}

Computational literary studies has established the value of large-scale evidence for literary history through distant reading, macroanalysis, and corpus-level stylistics \cite{moretti2013,jockers2013,underwood2019}. Network-oriented work has pushed the same conversation further by showing that literary organization often appears more clearly in relations than in counts alone, whether the units are characters, authors, or influence pathways \cite{elson2010,nazm2025,zhao2022}. Even so, the usual analytic units remain the token, the document, or the social actor. Symbolism poses a different problem. Its salient unit is often a culturally stabilized cluster of related forms.

The gap is especially visible in Persian literary computation. Recent work has improved corpus building, era or style classification, and exploratory AI-assisted interpretation \cite{raji2023,ruma2022,meymandi2024}. These studies make large-scale Persian analysis possible, but they usually model poems, poets, or lexical items rather than symbolic constellations. Persian literary scholarship has long treated imagery differently: as a historically sedimented system of figures whose meaning depends on reuse, adjacency, and convention rather than isolated lexical identity \cite{schimmel1992,seyedgohrab2011}. The question is not whether Persian poetry can be processed computationally. It is how to represent symbols without either fragmenting them into raw tokens or dissolving them into document-level themes.

This study also draws on diachronic semantics and dynamic network analysis. Work on semantic change has shown that historical meaning is often better captured by shifting relational profiles than by counts alone \cite{hamilton2016,momeni2018}. Dynamic-network research offers parallel ways to think about change through tie formation and dissolution, latent-space trajectories, community evolution, and diachronic node embeddings \cite{krivitsky2010,sewell2015,aggarwal2014,rossetti2018,xu2020,boccaletti2014}. Literary applications move in the same direction. Tanaka's phylogenetic modeling of waka and poet-network studies in Persian and Chinese traditions treat literary history as an evolving relational object rather than a static thematic inventory \cite{tanaka2025,nazm2025,zhao2022}. We extend that line of work by taking symbolic families, rather than single words or authors, as the basic analytic unit.

\section{Corpus and Method}

\subsection{Corpus and chronology}

The analysis is based on a corpus of 129{,}451 Persian poems attributed to 216 poets. After structural parsing, the corpus contains 1{,}446{,}226 verses and 2{,}891{,}826 hemistichs. Chronological analysis uses poet-level century metadata covering the full corpus and grouped into 11 Hijri-century bins, with the earliest sparse material merged into a single bin spanning the third and fourth centuries. Verse coverage under this mapping is complete for the analyzed corpus: every poem inherits a century from an explicit local poet-to-century table rather than from a language-only inference procedure. The chronology is therefore coarse, but its coarseness is visible and controlled rather than hidden.

\begin{table}[t]
\caption{Corpus and symbolic-inventory summary.}
\label{tab:corpus}
\centering
\small
\begin{tabular}{@{}lr@{}}
\toprule
Measure & Value \\
\midrule
Poems & 129{,}451 \\
Verses & 1{,}446{,}226 \\
Hemistichs & 2{,}891{,}826 \\
Poets & 216 \\
Century bins & 11 \\
Raw symbolic tokens consolidated & 1{,}160 \\
Canonical symbolic families & 738 \\
Families admitted to atlas construction & 175 \\
Unique families retained in the final atlas & 157 \\
\botrule
\end{tabular}
\end{table}

\subsection{Family induction and traceability}

The central methodological move is to treat the symbolic family, rather than the raw token, as the basic analytic unit. Families were built conservatively. Candidate merges were considered only for explicit surface relations such as clitic removal, prefix or suffix stripping, near-head extension, or synthetic heads supported by several strong variants. For each eligible merge, a composite score combined positive evidence from surface relation, static and occurrence-level embedding support, contextual overlap, cluster overlap, seed overlap, exemplar overlap, and head cleanliness, while subtracting penalties for fragment residue, function-word contamination, genericity noise, and proper-name inflation. In compact form,
\[
S_{\mathrm{merge}}=\sum_r w_r E_r-\sum_q \lambda_q P_q .
\]
Candidates were accepted only when the surface relation was eligible, the composite score cleared the staged threshold, and the merged head remained cleaner than its main variants. The staged schedule was conservative rather than expansive: strict early gates were relaxed only after a candidate had already passed the basic coherence checks, which is why the final inventory remains compact without becoming indiscriminate.

Traceability and head selection were enforced separately. Every raw token maps to exactly one family, and the token-to-family assignments remain inspectable in the preserved member and diagnostics tables. The canonical head is the root form of the accepted assignment chain; when no observed head is adequate, a synthetic head is admitted only if multiple variants jointly clear the synthetic-family threshold. In practice this keeps a family philologically legible even when several surface forms are absorbed.

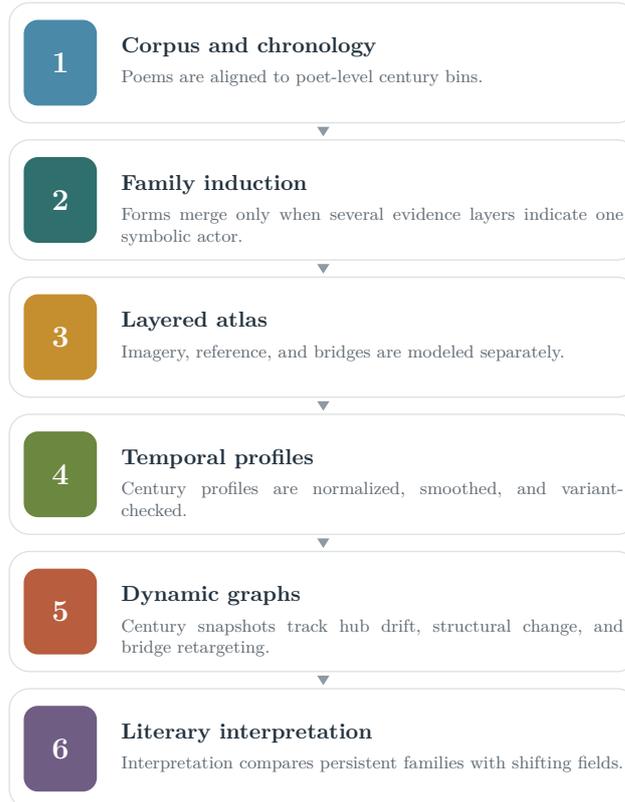
\begin{figure}[!htbp]
\centering
\resizebox{0.64\linewidth}{!}{%
\begin{tikzpicture}[x=1cm,y=1cm]
\tikzset{
  workflow box/.style={rounded corners=10pt, draw=workflowborder, fill=white, line width=0.65pt},
  workflow badge/.style={rounded corners=7pt, line width=0pt},
  workflow arrow/.style={-{Triangle[length=4.8pt,width=5.8pt]}, line width=1.1pt, draw=workflowarrow}
}
\draw[workflow box] (0.4,10.8) rectangle (11.3,8.7);
\fill[workflowsea, workflow badge] (0.65,10.5) rectangle (1.92,9.0);
\node[font=\bfseries\fontsize{15}{17}\selectfont, text=white] at (1.285,9.75) {1};
\node[anchor=north west, font=\bfseries\fontsize{10.8}{12.5}\selectfont, text=workflowink] at (2.22,10.32) {Corpus and chronology};
\node[anchor=north west, text width=8.72cm, align=justify, font=\fontsize{8.8}{10.7}\selectfont, text=workflowgray] at (2.22,9.74) {Poems are aligned to poet-level century bins.};

\draw[workflow box] (0.4,8.4) rectangle (11.3,6.3);
\fill[workflowteal, workflow badge] (0.65,8.1) rectangle (1.92,6.6);
\node[font=\bfseries\fontsize{15}{17}\selectfont, text=white] at (1.285,7.35) {2};
\node[anchor=north west, font=\bfseries\fontsize{10.8}{12.5}\selectfont, text=workflowink] at (2.22,7.92) {Family induction};
\node[anchor=north west, text width=8.72cm, align=justify, font=\fontsize{8.8}{10.7}\selectfont, text=workflowgray] at (2.22,7.34) {Forms merge only when several evidence layers indicate one symbolic actor.};

\draw[workflow box] (0.4,6.0) rectangle (11.3,3.9);
\fill[workflowgold, workflow badge] (0.65,5.7) rectangle (1.92,4.2);
\node[font=\bfseries\fontsize{15}{17}\selectfont, text=white] at (1.285,4.95) {3};
\node[anchor=north west, font=\bfseries\fontsize{10.8}{12.5}\selectfont, text=workflowink] at (2.22,5.52) {Layered atlas};
\node[anchor=north west, text width=8.72cm, align=justify, font=\fontsize{8.8}{10.7}\selectfont, text=workflowgray] at (2.22,4.94) {Imagery, reference, and bridges are modeled separately.};

\draw[workflow box] (0.4,3.6) rectangle (11.3,1.5);
\fill[workflowolive, workflow badge] (0.65,3.3) rectangle (1.92,1.8);
\node[font=\bfseries\fontsize{15}{17}\selectfont, text=white] at (1.285,2.55) {4};
\node[anchor=north west, font=\bfseries\fontsize{10.8}{12.5}\selectfont, text=workflowink] at (2.22,3.12) {Temporal profiles};
\node[anchor=north west, text width=8.72cm, align=justify, font=\fontsize{8.8}{10.7}\selectfont, text=workflowgray] at (2.22,2.54) {Century profiles are normalized, smoothed, and variant-checked.};

\draw[workflow box] (0.4,1.2) rectangle (11.3,-0.9);
\fill[workflowrust, workflow badge] (0.65,0.9) rectangle (1.92,-0.6);
\node[font=\bfseries\fontsize{15}{17}\selectfont, text=white] at (1.285,0.15) {5};
\node[anchor=north west, font=\bfseries\fontsize{10.8}{12.5}\selectfont, text=workflowink] at (2.22,0.72) {Dynamic graphs};
\node[anchor=north west, text width=8.72cm, align=justify, font=\fontsize{8.8}{10.7}\selectfont, text=workflowgray] at (2.22,0.14) {Century snapshots track hub drift, structural change, and bridge retargeting.};

\draw[workflow box] (0.4,-1.2) rectangle (11.3,-3.3);
\fill[workflowplum, workflow badge] (0.65,-1.5) rectangle (1.92,-3.0);
\node[font=\bfseries\fontsize{15}{17}\selectfont, text=white] at (1.285,-2.25) {6};
\node[anchor=north west, font=\bfseries\fontsize{10.8}{12.5}\selectfont, text=workflowink] at (2.22,-1.68) {Literary interpretation};
\node[anchor=north west, text width=8.72cm, align=justify, font=\fontsize{8.8}{10.7}\selectfont, text=workflowgray] at (2.22,-2.26) {Interpretation compares persistent families with shifting fields.};

\draw[workflow arrow] (5.85,8.58) -- (5.85,8.46);
\draw[workflow arrow] (5.85,6.18) -- (5.85,6.06);
\draw[workflow arrow] (5.85,3.78) -- (5.85,3.66);
\draw[workflow arrow] (5.85,1.38) -- (5.85,1.26);
\draw[workflow arrow] (5.85,-1.02) -- (5.85,-1.14);
\end{tikzpicture}
}
\caption{Analytical workflow from corpus alignment to dynamic graph analysis.}
\label{fig:pipeline}
\end{figure}

At this scale, the model stays close to literary usage. A family can absorb orthographic variation, clitic fusion, and near-duplicate heads while remaining easy to inspect. It also keeps apart categories that a looser semantic grouping would blur. Sacred titles, courtly names, and imagistic families can all carry symbolic charge, but they do different work in the corpus. We discussed the grouping scheme with a specialist in Persian literary studies as a plausibility check rather than a formal annotation exercise. Figure~\ref{fig:pipeline} sketches the sequence from family recovery to graph construction and historical analysis.

\subsection{Atlas construction, edge weights, and community labels}

The graph uses three connected components. The symbolic core contains families whose main force is imagistic or figurative. A separate referential component contains sacred and courtly families whose names remain structurally important but should not determine the same topology. Cross-layer links record recurrent attachment between the two. Each retained family receives one primary residency in either the core or the referential component, and cross-layer participation is then tracked separately. Assignment depends on symbolic strength, genericity risk, referential pressure, and bridge-anchor behavior rather than a manual whitelist.

Edges are likewise evidence-weighted rather than frequency-only. For each candidate pair $(i,j)$ in layer $\ell$, the atlas combines multiview embedding support, weighted context overlap, cluster overlap, representative-context overlap, verse and document co-association, poet overlap, shared seed support, and pairwise local co-activation into a single layer-specific score,
\[
S_{ij}^{(\ell)}=\sum_m w_m A_{ij}^{(m)}-\gamma_\ell G_{ij}+\rho_\ell R_{ij}.
\]
Here $G_{ij}$ is a genericity penalty applied in every component, while $R_{ij}$ captures referential pressure: it is negative in the core, near-neutral in the referential component, and slightly positive in the bridge view, where attachment to sacred or courtly material is part of the target phenomenon. After scoring, each component applies a fixed threshold and then a top-$k$ sparsification rule that becomes stricter as node genericity rises. The retained specification uses thresholds of $0.41$, $0.36$, and $0.38$ for the core, referential, and bridge components, with top-$k$ limits of $6$, $4$, and $4$ before node-specific genericity adjustment. The result is sparse enough to read in print. These choices were fixed from construction diagnostics before the diachronic analysis.

Community detection uses weighted greedy modularity on the core graph, both for the global network and for century-specific snapshots. Field labels come from overlap between community heads and anchor vocabularies for wine-tavern ritual, floral-vegetal, body-beloved, light-fire, water-sea, ascetic-mystical, and the two bridge domains. When overlap is weak and no seed anchor is present, the community remains in the mixed field. The labels are descriptive summaries of dominant membership, not stronger claims than the evidence supports.

\subsection{Diachronic and dynamic analysis}

Historical analysis proceeds in two steps. First, family frequencies are tracked across century bins after normalization by verse volume. The main series is reported per 10{,}000 verses, with a three-bin rolling mean and a Gaussian smoothing pass with $\sigma = 1.0$. To check poet concentration, we also compute poet-balanced weighting and top-poet-removal variants. Trend direction is evaluated from the smoothed trajectories with linear slope, Spearman monotonicity, and chi-square distribution-shift tests, then filtered through $100$ poet-bootstrap resamples. Families whose peak centuries or slope signs remain unstable under these checks are marked uncertain.

Second, century-specific graphs are derived from the same backbone so that structural change can be read within a common symbolic space. Dynamic nodes are activated by century-specific family presence; dynamic edges are activated by century-restricted co-activation on existing backbone edges. From these graphs we track hub drift, community change, and bridge retargeting.

The dynamic graph is rebuilt under five variants: loose, base, and strict activity thresholds, a poet-balanced weighting specification, and a high-confidence-only subset. In the base specification, families require at least $5$ raw occurrences, $3$ verse-presence events, and $2$ edge co-activations per century; the loose and strict variants shift those cutoffs to $3/2/1$ and $10/5/3$. We foreground claims that survive across these variants and label weaker ones as threshold-sensitive. Appendix~\ref{app:methods} lists the retained parameter values and robustness variants.

\section{Results}

\subsection{Recovering symbolic families from lexical scatter}

A token-only view fragments recurring symbolic material into partially overlapping forms and lets prestige vocabulary, generic reuse, and surface variation compete with imagistic material on the same footing. Family induction reduces that scatter without giving up lexical traceability. Because most families remain small, the consolidation step is selective rather than blanket lemmatization.

\begin{table}[!htbp]
\caption{Compact validation of family-level modeling.}
\label{tab:family_validation}
\centering
\footnotesize
\begin{tabularx}{\linewidth}{@{}p{0.43\linewidth}p{0.16\linewidth}X@{}}
\toprule
Diagnostic & Value & Summary \\
\midrule
Raw symbolic tokens $\rightarrow$ canonical families & $1{,}160 \rightarrow 738$ & Family induction removes $36.4\%$ of analytic scatter while preserving token-level traceability. \\
Median family size / single-member share & $1$ / $83.7\%$ & Consolidation is conservative rather than indiscriminate; most forms are left untouched unless evidence is strong. \\
Top-250 raw candidates $\rightarrow$ distinct families / clear fragment reduction & $250 \rightarrow 183$ / $98.65\%$ & The same window loses $67$ duplicate raw positions and absorbs $74$ clear fragment-like forms into stronger heads, producing cleaner rankings. \\
Seed-family recovery & $18/18$ & The manually seeded symbolic anchors are all recovered in the induced family inventory. \\
Referential leakage into top 40 symbolic-core families & $0$ & The induced core stays imagistic rather than being overrun by sacred or courtly name-like material. \\
\botrule
\end{tabularx}
\end{table}

Table~\ref{tab:family_validation} shows the immediate effect of working at this scale. The inventory becomes smaller and cleaner, yet most families remain singletons. In the top-250 raw window, repeated surface variants stop taking multiple ranking slots, and sacred or courtly material stays out of the symbolic core.

\subsection{Symbolic structure: imagery, reference, and attachment}

In the retained graph, the symbolic core is relatively sparse, the referential component much denser, and the attachment zone between them restricted rather than diffuse (Table~\ref{tab:layers}; Figures~2A--2C). The unique retained inventory contains $157$ families assigned primarily to the core or referential component; bridge participation is an overlapping view on that same set. Imagistic symbols form selective neighborhoods. Sacred and courtly names cluster more tightly among themselves, and only some of them attach recurrently to the core.

\begin{table}[!htbp]
\caption{Structural summary of the symbolic graph.}
\label{tab:layers}
\centering
\footnotesize
\begin{tabular}{@{}lrrrrp{0.31\linewidth}@{}}
\toprule
Layer & Nodes in layer & Edges & Avg. degree & Density & Role \\
\midrule
Imagistic core & 127 & 223 & 3.51 & 0.028 & Recurrent symbolic fields built from floral, bodily, ritual, elemental, and spatial imagery \\
Referential layer & 30 & 81 & 5.40 & 0.186 & Sacred and courtly names treated as meaningful reference rather than folded into imagistic symbolism \\
Bridge layer & 63 & 143 & 4.54 & 0.073 & Attachment zone linking referential materials to symbolic neighborhoods \\
\addlinespace[2pt]
\multicolumn{6}{p{\linewidth}}{\footnotesize Note: The bridge row is a participation count rather than an additional disjoint node inventory. The unique atlas total in Table~\ref{tab:corpus} reports the 157 retained families assigned primarily to the core or referential residencies; the bridge count records how many of those same retained families also participate in cross-layer attachment edges.} \\
\botrule
\end{tabular}
\end{table}

Within the core, the dominant communities are recognizably literary.

\begin{figure}[!htbp]
\centering
\includegraphics[width=0.76\linewidth]{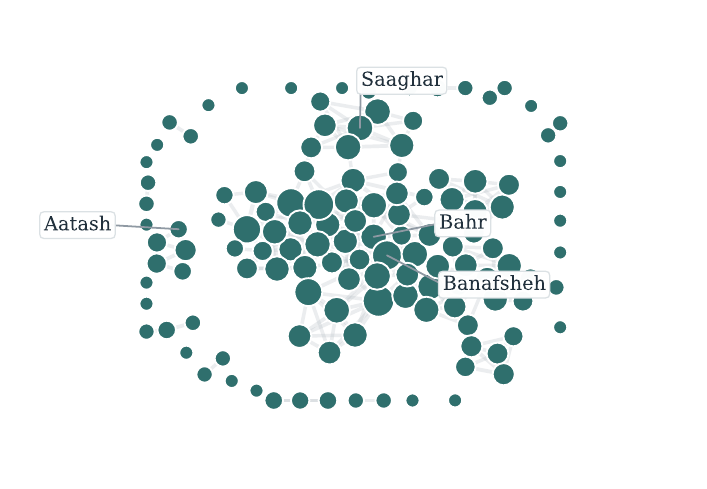}
\manualfigcaption{\textbf{Figure 2A.}\hskip.7em Imagistic Core. The sparse symbolic core organizes recurrent families such as \symgloss{Saaghar}{Wine Cup}, \symgloss{Aatash}{Fire}, and \symgloss{Bahr}{Sea} into legible imagistic neighborhoods.}
\end{figure}

\begin{figure}[!htbp]
\centering
\includegraphics[width=0.76\linewidth]{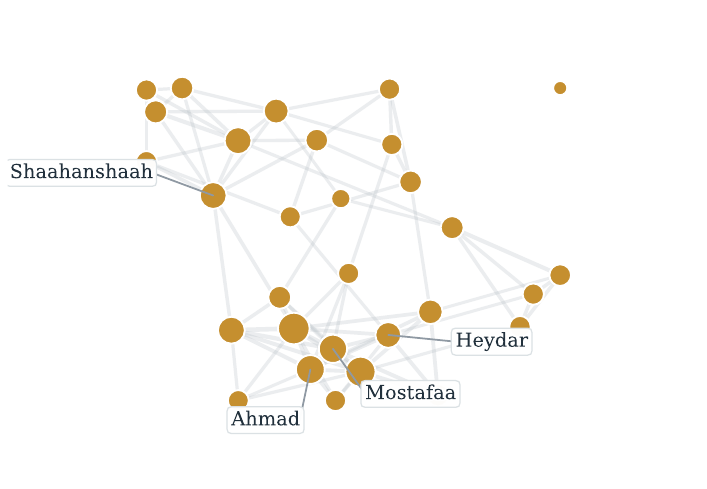}
\manualfigcaption{\textbf{Figure 2B.}\hskip.7em Referential Layer. Sacred and courtly reference forms a denser cluster around titles such as \symgloss{Payghambar}{Prophet} and royal designations such as \symgloss{Shaahanshaah}{King of Kings}.}
\end{figure}

\begin{figure}[!htbp]
\centering
\includegraphics[width=0.76\linewidth]{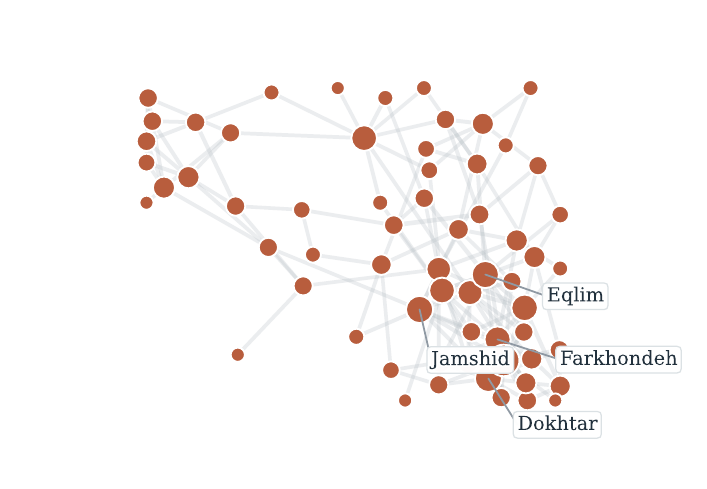}
\manualfigcaption{\textbf{Figure 2C.}\hskip.7em Bridge Layer. Mediating families such as \symgloss{Eqlim}{Realm}, \symgloss{Farkhondeh}{Blessed}, and \symgloss{Dokhtar}{Daughter} carry reference toward imagistic fields without collapsing the two layers.}
\end{figure}

One cluster joins tavern, wine vessel, and ritual language; another gathers bodily beauty, hair, and fragrance; others organize around sea imagery, vegetal verticality, and light or fire. These are coordinated fields rather than isolated emblematic words. Figure~3 makes that visible at a glance.

The referential component is not a residue left outside the core. Sacred titles and courtly names remain active and historically meaningful, but they behave differently from imagistic families. Their strongest recurring attachments concentrate on the mixed field, where prestige, devotion, and imagery meet through recurrent contact zones.

\subsection{Stable anchors, rising families, and declining prestige vocabularies}

Century-level trajectories show a system that changes without losing its backbone. After smoothing and robustness filtering, 29 families are robustly rising, 9 robustly declining, and 4 stable across the full chronology. Continuity is present, but not as stasis. Persistent families remain in place while the surrounding fields are reweighted, and the bootstrap peak shares in Table~\ref{tab:trajectories} keep the remaining temporal uncertainty explicit.

\begin{figure}[t]
\centering
\includegraphics[width=0.76\linewidth]{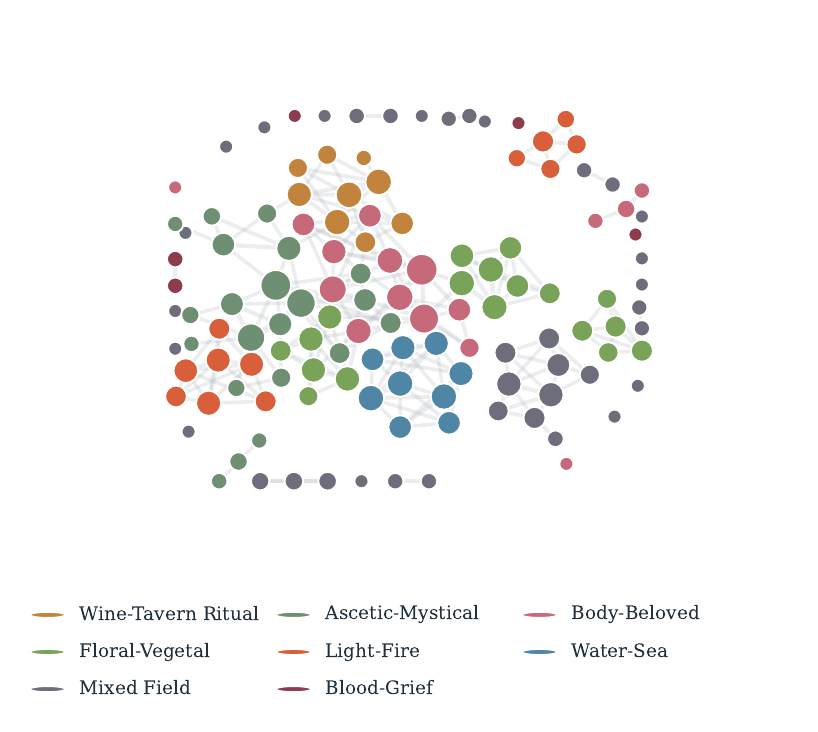}
\manualfigcaption{\textbf{Figure 3.}\hskip.7em Community structure in the imagistic core. Colors identify the principal symbolic fields recovered from the atlas without further in-figure annotation.}
\end{figure}

The rising families concentrate in a late-strengthening lyric ecology. \symfirst{پیمانه}{Peymaaneh}{Wine Vessel} and \symfirst{صهبا}{Sahbaa}{Wine} rise alongside families tied to gardens, flame, melody, and sensorial surface, including \symfirst{گلشن}{Golshan}{Garden}. This is not only a matter of raw frequency. The same late centuries also show tighter coordination among the wine-tavern, floral, and body-beloved fields. The strongest field-level coupling is between floral-vegetal and wine-tavern ritual (smoothed correlation $= 0.983$), followed closely by body-beloved with wine-tavern ritual ($0.982$) and body-beloved with light-fire ($0.979$).

The declining families point in another direction. \symfirst{خاقان}{Khaaqaan}{Sovereign}, \symfirst{فرخ}{Farrokh}{Auspicious}, \symfirst{باد}{Baad}{Wind}, and related items are earlier-concentrated and then recede. What declines here is not ``reference'' in general, but a particular prestige-coded register associated with courtly and heroic diction. The pattern does not justify a simplistic historical opposition between ``courtly'' and ``mystical'' poetry. Even so, the redistribution is clear enough to matter: earlier prestige vocabulary loses relative weight while later lyric-sensorial families become more central to the symbolic profile.

Stable anchors push against overdramatic periodization. \symfirst{شب}{Shab}{Night}, \symfirst{روز}{Ruz}{Day}, and \symfirst{خاک}{Khaak}{Earth} remain distributed across the chronology with high temporal entropy and low drift. They are not specialized motifs tied to one historical moment. They function more like durable coordinates around which other families rise or fall.

\begin{table}[t]
\caption{Representative family trajectories after temporal refinement. Peak-mode share reports the proportion of $100$ poet-bootstrap resamples that recover the same modal peak century.}
\label{tab:trajectories}
\centering
\footnotesize
\setlength{\tabcolsep}{3.5pt}
\renewcommand{\arraystretch}{1.02}
\begin{tabularx}{\linewidth}{@{}l l c c l X@{}}
\toprule
Family & Trajectory & Peak & Peak-mode share & Field & Pattern \\
\midrule
\symgloss{Peymaaneh}{Wine Vessel} & Robust rising & 12 & 76\% & Wine-tavern ritual & Late strengthening of vessel imagery within lyric-tavern clusters \\
\symgloss{Sahbaa}{Wine} & Robust rising & 14 & 78\% & Wine-tavern ritual & Sustained late growth of wine nomenclature beyond a single century spike \\
\symgloss{Golshan}{Garden} & Robust rising & 12 & 81\% & Floral-vegetal & Expansion of garden space as a late symbolic environment \\
\symgloss{Khaaqaan}{Sovereign} & Robust declining & 6 & 63\% & Referential / courtly & Recession of prestige-coded royal diction after early prominence \\
\symgloss{Farrokh}{Auspicious} & Robust declining & 5 & 93\% & Mixed field & Weakening of auspicious courtly language inside the broader contact zone \\
\symgloss{Baad}{Wind} & Robust declining & 6 & 67\% & Mixed field & Earlier-heavy mobility and force imagery loses weight historically \\
\symgloss{Shab}{Night} & Stable timeless & 12 & 51\% & Mixed field & Persistent lyric anchor with broad distribution across the chronology \\
\symgloss{Khaak}{Earth} & Stable timeless & 11 & 90\% & Mixed field & Durable ethical and metaphysical anchor rather than a period-bound image \\
\botrule
\end{tabularx}
\end{table}

Figures~4A--4C show the contrast directly. The stable curves remain present across nearly the whole chronology, while the rising and declining curves track redistribution within the same repertoire.

\begin{figure}[H]
\centering
\includegraphics[width=0.70\linewidth]{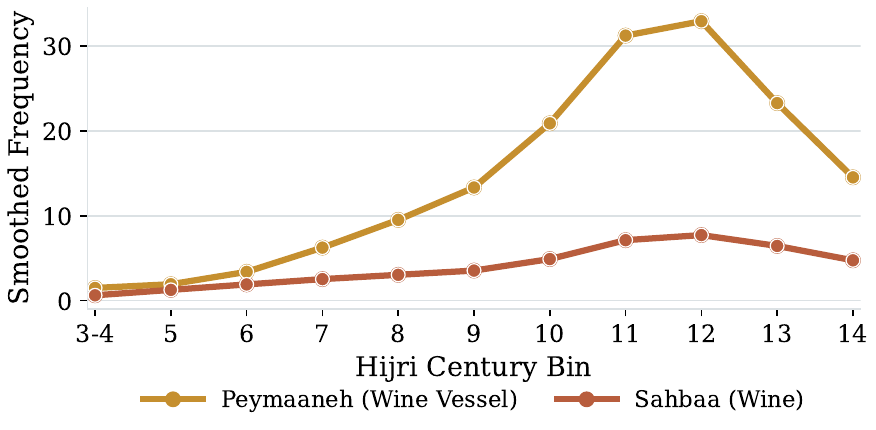}
\manualfigcaption{\textbf{Figure 4A.}\hskip.7em Later-Strengthening Families. \symgloss{Peymaaneh}{Wine Vessel} and \symgloss{Sahbaa}{Wine} intensify most strongly in the later chronology.}
\end{figure}

\begin{figure}[H]
\centering
\includegraphics[width=0.70\linewidth]{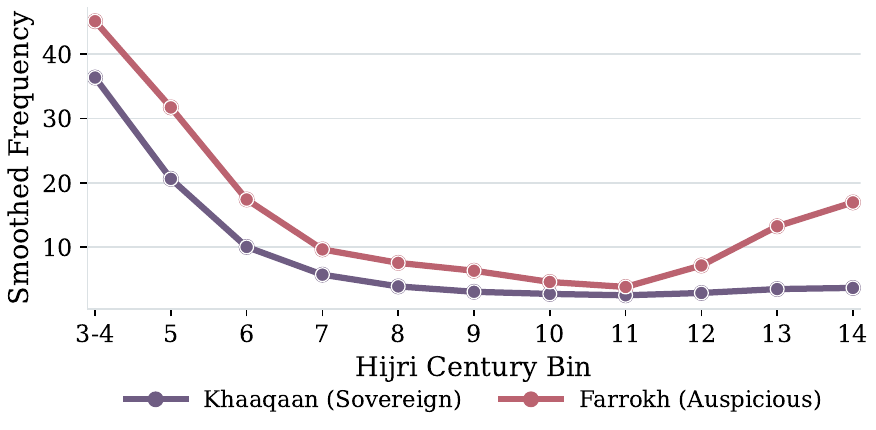}
\manualfigcaption{\textbf{Figure 4B.}\hskip.7em Earlier-Concentrated Families. \symgloss{Khaaqaan}{Sovereign} and \symgloss{Farrokh}{Auspicious} retain their strongest weight in the earlier bins.}
\end{figure}

\begin{figure}[H]
\centering
\includegraphics[width=0.70\linewidth]{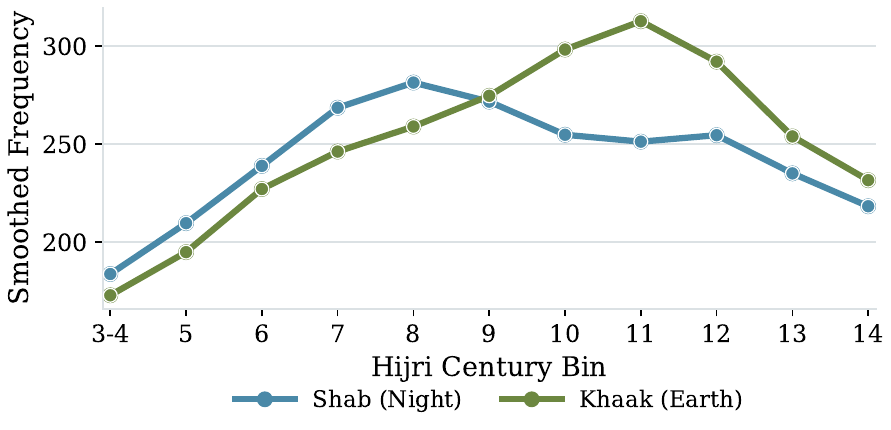}
\manualfigcaption{\textbf{Figure 4C.}\hskip.7em Stable Anchors. \symgloss{Shab}{Night} and \symgloss{Khaak}{Earth} remain broadly distributed across the mapped chronology.}
\end{figure}

\subsection{Structural change across centuries}

The historical signal becomes clearer when the analysis shifts from family frequencies to network structure. Century-specific graphs show a system that stays recognizable while changing its internal arrangement. The late-minus-early deltas are modest, but they recur in the same direction often enough to keep. Core modularity rises in the base specification by $+0.020$, and the same direction appears in four of the five robustness variants. Cross-scope linkage declines by $-0.032$ in the base specification, again with the same direction in four of five variants. Degree centralization is essentially flat. What changes, then, is not the existence of the network but the separation of its fields.

Hub drift makes that reorganization tangible. \symfirst{خرقه}{Kherqe}{Sufi Robe} is the clearest rising hub, reaching its highest structural prominence late in the chronology. \symfirst{فرخنده}{Farkhondeh}{Blessed}, \symfirst{بنفشه}{Banafsheh}{Violet}, and related early-centered families are much stronger in earlier and middle centuries than they are late. \symfirst{ساغر}{Saaghar}{Wine Cup} remains a stable hub throughout, and stable hubs such as \symfirst{باده}{Baadeh}{Wine} and \symfirst{گل}{Gol}{Rose} reinforce the same point: the symbolic repertoire persists even while the surrounding network changes. What shifts, then, is not simply which families exist, but which ones organize the symbolic space at a given historical moment.

\begin{table}[t]
\caption{Dynamic network changes with robustness and uncertainty summary.}
\label{tab:rewiring}
\centering
\footnotesize
\begin{tabularx}{\linewidth}{@{}p{0.22\linewidth}p{0.10\linewidth}p{0.23\linewidth}p{0.13\linewidth}X@{}}
\toprule
Metric & Base change & Uncertainty summary & Variant support & Interpretation \\
\midrule
Core modularity (late minus early) & $+0.020$ & Comparable-variant span $+0.012$ to $+0.030$; permutation upper-tail $0.14$ & Positive in $4/5$ variants & Small but fairly consistent separation among fields. \\
Cross-link ratio (late minus early) & $-0.032$ & Comparable-variant span $-0.032$ to $-0.019$; permutation lower-tail $0.05$ & Negative in $4/5$ variants & Cross-scope links decline across most reconstructions. \\
Courtly bridge strength (late minus early) & $-0.033$ & Variant span $-0.033$ to $-0.007$ & Negative in $2/4$, flat in $2/4$ & Courtly weakening is visible, though it depends more on threshold choice. \\
Sacred bridge strength (late minus early) & $+0.049$ & Variant span $+0.042$ to $+0.071$ & Positive in $4/4$ variants & Sacred-prophetic attachment strengthens late with the strongest support in this table. \\
Degree centralization (late minus early) & $-0.001$ & Comparable-variant span $-0.002$ to $-0.001$; high-confidence subset $+0.037$ & Flat in $4/5$ variants & Centralization contributes little to the main historical contrast. \\
\botrule
\end{tabularx}
\end{table}

Community-level change points in the same direction. The wine-tavern ritual field is the most persistent major community, with a mean active-member ratio of $0.990$. Light-fire and water-sea are also highly stable. Floral-vegetal stays active but has the highest split pressure ($0.392$), so it changes internally more than the other large lyric fields. Body-beloved shows stronger merge pressure toward floral-vegetal, which indicates recurrent traffic between bodily and garden imagery. Ascetic-mystical is smaller, but its strongest external pull is toward wine-tavern ritual.

Figures~5A--5B and 6A--6B plot the same shifts. Modularity rises gradually, cross-scope linkage falls, and families do not move in parallel. Courtly bridges weaken overall, whereas sacred-prophetic bridges gain strength late while remaining centered on the mixed field.

\begin{figure}[H]
\centering
\includegraphics[width=0.70\linewidth]{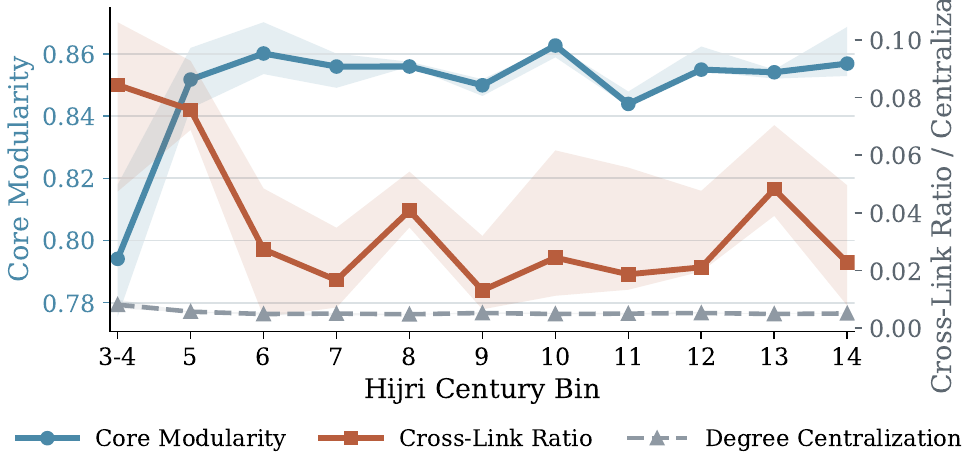}
\manualfigcaption{\textbf{Figure 5A.}\hskip.7em Topological Evolution. Core modularity rises over time while cross-scope linkage declines and degree centralization remains comparatively flat. Shaded ribbons show the span across the loose, base, strict, and poet-balanced reconstructions.}
\end{figure}

\begin{figure}[H]
\centering
\includegraphics[width=0.70\linewidth]{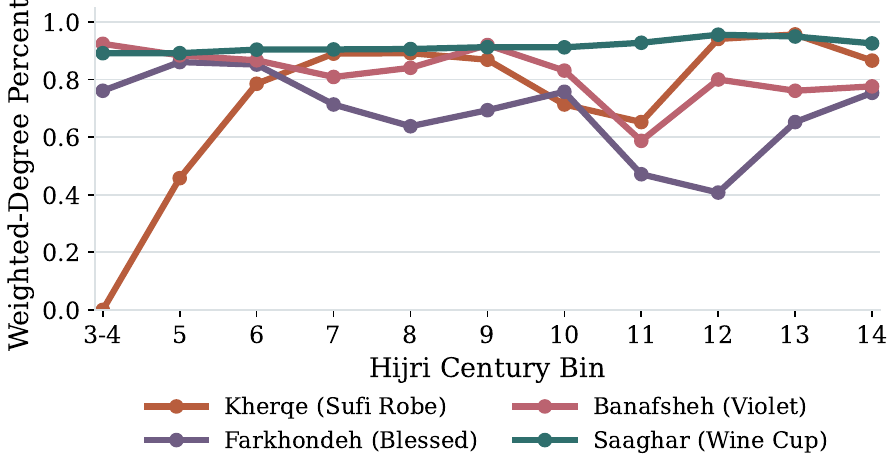}
\manualfigcaption{\textbf{Figure 5B.}\hskip.7em Hub Drift. \symgloss{Kherqe}{Sufi Robe} strengthens late, whereas \symgloss{Farkhondeh}{Blessed} and \symgloss{Banafsheh}{Violet} lose structural prominence; \symgloss{Saaghar}{Wine Cup} remains comparatively stable.}
\end{figure}

\begin{figure}[H]
\centering
\includegraphics[width=0.70\linewidth]{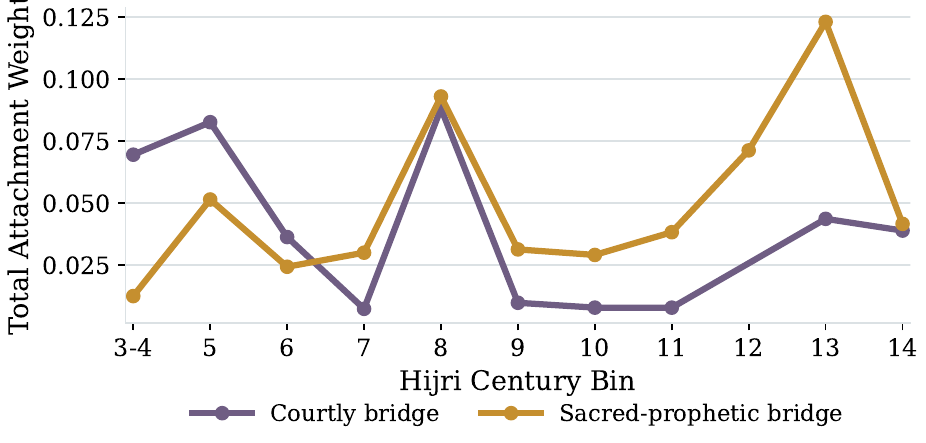}
\manualfigcaption{\textbf{Figure 6A.}\hskip.7em Attachment Strength. The \textit{Sacred-Prophetic Bridge} strengthens overall, whereas the \textit{Courtly Bridge} weakens or remains threshold-sensitive.}
\end{figure}

\begin{figure}[H]
\centering
\includegraphics[width=0.70\linewidth]{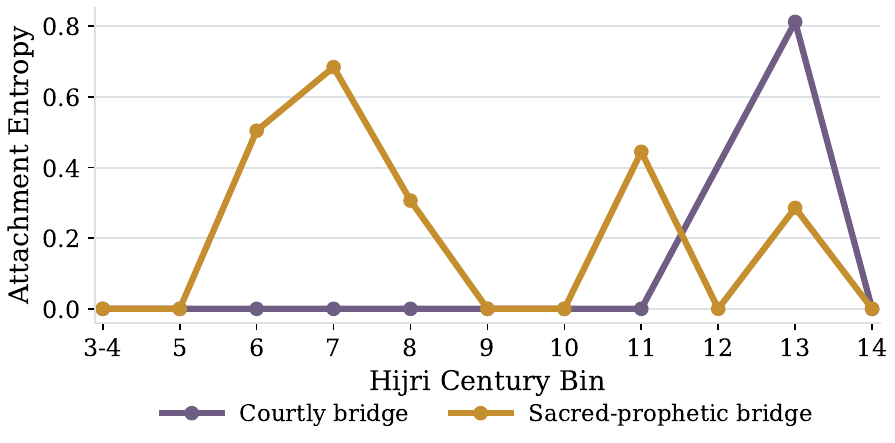}
\manualfigcaption{\textbf{Figure 6B.}\hskip.7em Attachment Dispersion. The two bridge domains show uneven but differentiating patterns of target dispersion across the chronology.}
\end{figure}

\subsection{Poet-level differentiation, clustering, and historical breadth}

We also recalculated the field profiles at poet level. The threshold-based core pool contains $36$ poets with at least $10{,}000$ verses and $1{,}000$ symbolic-family occurrences. The rule is coverage-driven rather than prestige-driven: it includes Saadi, Khaaqani, and Vahshi Bafqi under the same standard, while Hafez falls below the mapped verse floor and is excluded from both the core pool and the broader sensitivity check. Figure~7 orders the eligible poets by Ward clustering on standardized field profiles; Table~\ref{tab:poet_affinities} gives the strongest field-level concentrations, and Appendix Tables~\ref{tab:appendix_poet_profiles}--\ref{tab:appendix_poet_neighbors} provide the full poet statistics.

The poet space is structured rather than noisy. A three-cluster solution yields the clearest literary partition: a \textit{lyric-amplifying} cluster with high wine, body, floral, and mixed-field rates; a \textit{sparse-anchor} cluster with lower-than-average field intensity across the lyric system; and a broader \textit{mediating-balanced} cluster that contains poets such as Sana'i, Khaaqani, Saadi, Mowlana, Vahshi Bafqi, and Bidel Dehlavi. The clusters are not reducible to century bins. The lyric-amplifying group stretches from the seventh to the thirteenth centuries; the sparse-anchor group joins fourth-, fifth-, sixth-, thirteenth-, and fourteenth-century poets; and the mediating cluster spans the fifth through thirteenth centuries. Chronology matters, but it does not exhaust poet-level symbolic differentiation.

Similarity structure sharpens that point. In standardized field space, Khaaqani and Vahshi Bafqi remain comparatively central rather than isolated, whereas Saeb, Jahan Malek Khatun, and Ashofteh Shirazi are the clearest outliers. Neighbor relations also cross chronology: Saadi lies closest to Sana'i and Anvari, while Khaaqani sits near Mowlana rather than only beside adjacent sixth-century poets. Shared tradition survives as patterned resemblance, but poets amplify it unevenly and sometimes along noncontiguous historical lines.

\begin{figure}[!t]
\centering
\includegraphics[width=0.98\linewidth]{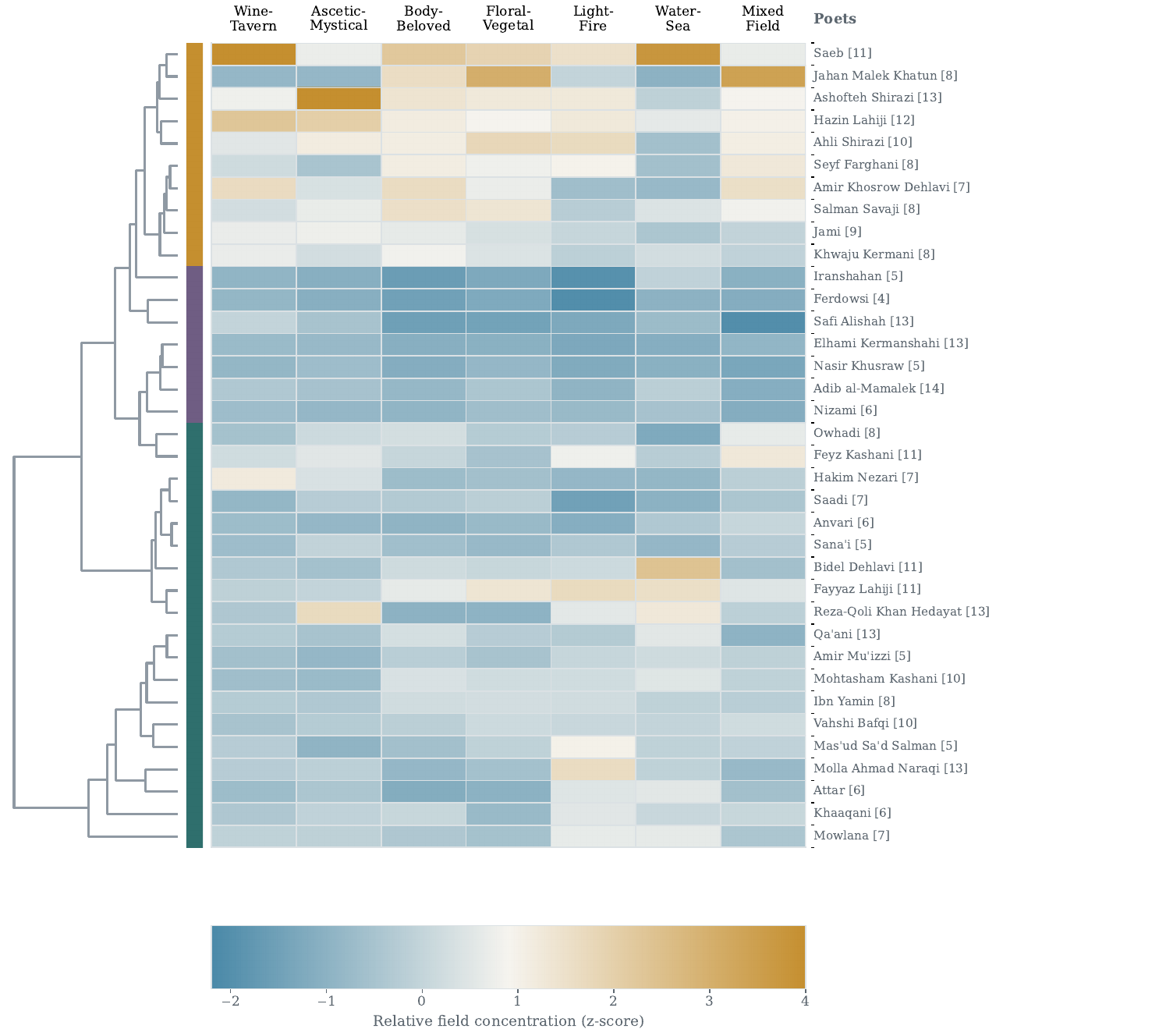}
\manualfigcaption{\textbf{Figure 7.}\hskip.7em Clustered poet-field profiles. Eligible poets are ordered by Ward clustering on verse-normalized field rates; the side strip marks the three retained clusters, with gold for lyric-amplifying, teal for mediating-balanced, and plum for sparse-anchor profiles. Warmer cells indicate stronger-than-average concentration within a field, cooler cells weaker-than-average concentration.}
\end{figure}

The poet comparison also helps with the century effect. On a broader sensitivity pool of $58$ poets with at least $8{,}000$ verses and $500$ symbolic-family occurrences, $24/27$ later poets lie above the earlier median for \textit{Wine-Tavern Ritual}, $20/27$ for \textit{Floral-Vegetal}, $21/27$ for \textit{Body-Beloved}, and $25/27$ for \textit{Ascetic-Mystical}. Those shifts persist after the single strongest late poet is removed: the late median remains above the earlier median in wine ($580.6$ versus $292.6$), floral ($679.8$ versus $458.0$), body ($859.2$ versus $585.5$), and ascetic-mystical language ($117.6$ versus $44.1$). Saeb alone supplies $45.7\%$ of late wine-tavern occurrences, yet the field still rises across nearly the whole later pool.

Table~\ref{tab:poet_affinities} shows the strongest concentrations inside that shared system. Saeb concentrates wine, body, and water imagery; Jahan Malek Khatun intensifies floral and mixed fields; Ahli Shirazi peaks in light-fire; and Ashofteh Shirazi in ascetic-mystical language. The repertoire is shared, but poets redistribute emphasis unevenly within it.

\begin{table}[!t]
\caption{Leading poet-level field affinities in the threshold-based core poet pool. Rates report field occurrences per $10{,}000$ verses among mapped poets with at least $10{,}000$ verses and $1{,}000$ symbolic-family occurrences.}
\label{tab:poet_affinities}
\centering
\footnotesize
\begin{tabularx}{\linewidth}{@{}p{0.20\linewidth}p{0.20\linewidth}p{0.13\linewidth}X@{}}
\toprule
Field & Leading poet & Rate / $10{,}000$ verses & Characteristic family heads \\
\midrule
Wine-Tavern Ritual & Saeb (11th c.) & 3250.2 & \symgloss{Mey}{Wine}; \symgloss{Baadeh}{Wine}; \symgloss{Sharaab}{Wine} \\
Ascetic-Mystical & Ashofteh Shirazi (13th c.) & 604.7 & \symgloss{Meykhaaneh}{Wine House}; \symgloss{Meykadeh}{Wine House}; \symgloss{Moghaan}{Magi} \\
Body-Beloved & Saeb (11th c.) & 1624.6 & \symgloss{Cheshm}{Eye}; \symgloss{Lab}{Lip}; \symgloss{La'l}{Ruby} \\
Floral-Vegetal & Jahan Malek Khatun (8th c.) & 1770.9 & \symgloss{Sarv}{Cypress}; \symgloss{Gol}{Rose}; \symgloss{Qad}{Stature} \\
Light-Fire & Ahli Shirazi (10th c.) & 692.2 & \symgloss{Aatash}{Fire}; \symgloss{Nur}{Light}; \symgloss{Zolmat}{Darkness} \\
Water-Sea & Saeb (11th c.) & 771.0 & \symgloss{Daryaa}{Sea}; \symgloss{Mowj}{Wave}; \symgloss{Bahr}{Sea} \\
Mixed Field & Jahan Malek Khatun (8th c.) & 6457.3 & \symgloss{Del}{Heart}; \symgloss{Eshq}{Love}; \symgloss{Shab}{Night} \\
\botrule
\end{tabularx}
\end{table}

\subsection{Literary context for the symbolic shifts}

The poet-level contrasts are easier to place once they are read back into literary history. Saadi and Mowlana occupy the broad mediating cluster rather than the most amplified lyric pole. That fits their historical role in consolidating the ghazal and didactic-mystical lyric as flexible symbolic forms rather than narrowly specialized repertoires. Saadi keeps bodily and floral material in measured relation to a stable mixed field. Mowlana leans more strongly toward light-fire and water-sea imagery, which suits an ecstatic symbolic economy in which motion, illumination, and dissolution matter more than courtly designation.

Khaaqani points in a different direction. Although strongly associated with learned panegyric and courtly rhetoric, he remains symbolically central rather than sealed off in a prestige enclave. What declines over time is not rhetorical density as such, but the sustained weight of royal and prestige-coded symbolic heads. Courtly discourse loses relative force even while highly wrought poets continue to share the larger imagistic network.

The later lyric shift is sharper. Saeb's concentration in wine, body, and water fields matches the intensified imagistic chaining usually associated with later ghazal practice and, more specifically, with the dense figurative pressure of \textit{sabk-e hendi}. Bidel belongs to the same later chronology without reproducing Saeb's extremity: his profile stays more mediating, consistent with a more abstract and reflective late style. Jahan Malek Khatun and Ahli Shirazi show that amplification is field-specific rather than uniform, with one poet intensifying floral-mixed continuity and the other light-fire concentration. In this view, the rise of wine-tavern and adjacent lyric fields is not just an increase in sensual vocabulary. It marks a broader reorganization in which mystical metaphor, bodily lyricism, and imagistic density become more tightly coupled, while explicitly courtly symbolism recedes.

\FloatBarrier

\section{Discussion}

The main pattern is continuity with rearrangement. Night, day, and earth remain widely distributed, while later centuries bring stronger coordination among wine, garden, body, and light. Earlier prestige-coded and heroic vocabularies lose relative ground. The raw effect sizes are often modest, but the direction of change repeats across variants and across family-, field-, and graph-level summaries.

Reference changes in a different way. Sacred figures and royal names remain symbolically consequential, but they do not behave like ordinary imagistic families. Keeping them in a separate component shows where they attach and where they do not. The mixed field matters here because it absorbs a large share of those recurring attachments.

Methodologically, the family scale sits between token and document. It stays close to the lexicon while still supporting historical comparison. Persian symbolism is one case where that middle unit is useful, but the same logic could apply to motifs, ritual vocabularies, epithets, or recurring imagistic frames in other traditions.

For Persian literary history, the result is fairly simple. The tradition does not reduce to a timeless storehouse of emblems, and it does not dissolve into a sequence of disconnected lexical fashions. Familiar resources remain available, but their relations are historically rearranged.

\section{Limitations}

Several limitations shape these claims. First, the chronology is poet-level rather than poem-level, so internal career development, uncertain dating, and later textual transmission are compressed into coarse century bins. Second, the family inventory remains an interpretive construction even when it is strongly constrained by contextual evidence. Some border cases, especially in the mixed field, still admit alternative classification.

Third, the dynamic graphs are derived from a common backbone. That improves comparability across centuries, but it also means that century-specific graphs are changing subsets of a shared symbolic space rather than independently discovered networks. Fourth, the null evidence remains modest: the chronology-order permutation baseline tests whether the observed directional claims are stronger than arbitrary reordering of the stored snapshots, but it is not a full generative null over poems or edges. Not every topological signal is equally strong. Modularity and cross-scope linkage move more consistently than centralization, and courtly bridge weakening is more threshold-sensitive than sacred bridge strengthening.

\section{Conclusion}

Modeling symbolic families changes the view of Persian poetry at scale. Grouped forms produce a network that is sparse in its imagistic center, denser in reference, and shaped by recurring attachments between the two.

Across centuries, some families stay widely distributed, others rise or fade, and the network shifts incrementally. Modularity increases, cross-scope links decline, sacred bridges strengthen, and hub positions move. The change is relational more than inventory-wide.

\backmatter

\bmhead{Acknowledgements}

We thank the Ganjoor project for making broad digital access to Persian poetry possible and for enabling corpus-based work of this kind. We also acknowledge the Persian poetic tradition on which this study depends and approach that literary heritage with respect for its cultural importance.

\clearpage
\appendix

\section{Supplementary Appendix}\label{app:methods}

The supplementary materials provide compact support tables and a broader family inventory. They are meant to keep the symbolic inventory legible without repeating low-level variant detail.

\subsection{Scoring, uncertainty, and consultation summary}

Two points are worth stating directly. The unique atlas total reported in Table~\ref{tab:corpus} is $157$ because graph gating assigns each retained family to one primary residency: $127$ in the imagistic core and $30$ in the referential layer. The bridge layer is an overlapping participation view on that same set, not an additional disjoint inventory. We also discussed the family-level framing with a specialist in Persian literary studies as a plausibility check rather than a large formal annotation campaign.

\begin{table}[!t]
\caption{Compact scoring summary for family induction and edge construction.}
\label{tab:appendix_scoring}
\centering
\footnotesize
\begin{tabularx}{\linewidth}{@{}p{0.17\linewidth}p{0.29\linewidth}p{0.24\linewidth}X@{}}
\toprule
Stage & Positive evidence & Penalties and gates & Output rule \\
\midrule
Family induction & Eligible surface relation; static and occurrence-level embeddings; contextual and cluster overlap; seed and exemplar support; head cleanliness & Fragment residue, function-word contamination, genericity noise, and proper-name inflation lower the merge score; staged thresholds keep weak cases as singletons & Each token maps to one family; accepted heads default to the cleanest root form, with synthetic heads admitted only when several strong variants jointly clear the synthetic threshold \\
Edge construction & Multiview embedding support; weighted context overlap; cluster overlap; representative-context overlap; verse/document co-association; poet overlap; seed support; local co-activation & Genericity is penalized in every layer; referentiality is negative in the core, near-neutral in the referential layer, and slightly positive in the bridge layer; layer thresholds and top-$k$ sparsification prune generic neighborhoods more aggressively & Each retained family keeps one primary atlas residency, while bridge participation records which retained families also sustain cross-layer attachment edges \\
\botrule
\end{tabularx}
\end{table}

\begingroup
\footnotesize
\setlength{\tabcolsep}{3.2pt}
\begin{longtable}{@{}p{0.24\linewidth}p{0.12\linewidth}p{0.26\linewidth}p{0.30\linewidth}@{}}
\caption{Supplementary uncertainty summary for representative families and dynamic claims.}\label{tab:appendix_uncertainty}\\
\toprule
Target & Point estimate & Uncertainty summary & Summary \\
\midrule
\endfirsthead
\toprule
Target & Point estimate & Uncertainty summary & Summary \\
\midrule
\endhead
\symgloss{Peymaaneh}{Wine Vessel} & Slope $+2.51$ & $95\%$ bootstrap CI $[+1.37,+3.40]$; peak-share $76\%$ & Late strengthening remains stable under poet-bootstrap resampling. \\
\symgloss{Sahbaa}{Wine} & Slope $+0.60$ & $95\%$ bootstrap CI $[+0.29,+0.92]$; peak-share $78\%$ & The rise is smaller than \symgloss{Peymaaneh}{Wine Vessel} but remains consistent. \\
\symgloss{Golshan}{Garden} & Slope $+5.06$ & $95\%$ bootstrap CI $[+3.53,+6.82]$; peak-share $81\%$ & Garden-space intensification is one of the clearest late family gains. \\
\symgloss{Khaaqaan}{Sovereign} & Slope $-1.74$ & $95\%$ bootstrap CI $[-2.92,-0.14]$; peak-share $63\%$ & Early concentration is consistent, even if the exact peak century is not knife-edged. \\
\symgloss{Farrokh}{Auspicious} & Slope $-2.06$ & $95\%$ bootstrap CI $[-3.06,-0.39]$; peak-share $93\%$ & Prestige-coded decline is unusually stable among the declining families. \\
\symgloss{Baad}{Wind} & Slope $-19.31$ & $95\%$ bootstrap CI $[-27.97,-11.27]$; peak-share $67\%$ & Mobility imagery is earlier-weighted with a strongly negative slope. \\
\symgloss{Shab}{Night} & Slope $+2.09$ & $95\%$ bootstrap CI $[-2.51,+7.24]$; peak-share $51\%$ & The family is temporally diffuse rather than sharply periodized, so it is treated as stable rather than monotonic. \\
\symgloss{Khaak}{Earth} & Slope $+7.03$ & $95\%$ bootstrap CI $[+2.08,+11.20]$; peak-share $90\%$ & Later weighting is real, but the family remains distributed across the whole chronology and still functions as an anchor. \\
Core modularity & $\Delta +0.020$ & Comparable-variant span $+0.012$ to $+0.030$; permutation upper-tail $0.14$ & A modest but consistently positive structural separation signal. \\
Cross-link ratio & $\Delta -0.032$ & Comparable-variant span $-0.032$ to $-0.019$; permutation lower-tail $0.05$ & The decline repeats across the main reconstructions. \\
Courtly bridge strength & $\Delta -0.033$ & Variant span $-0.033$ to $-0.007$; flat in $2/4$ variants & Courtly weakening is visible but should be read cautiously. \\
Sacred bridge strength & $\Delta +0.049$ & Variant span $+0.042$ to $+0.071$ & Sacred-prophetic strengthening is the most secure bridge-level shift. \\
\bottomrule
\end{longtable}
\par\endgroup

\subsection{Supplementary family inventory}

Table~\ref{tab:appendix_inventory} gathers a larger set of representative families across the principal symbolic fields. The selection is not exhaustive. It is meant to show recognizable Persian forms, transliterations, and English glosses while keeping the family-level scale intact.

\begingroup
\footnotesize
\setlength{\tabcolsep}{3.2pt}
\begin{longtable}{@{}p{0.23\linewidth}p{0.18\linewidth}p{0.13\linewidth}p{0.38\linewidth}@{}}
\caption{Supplementary inventory of representative symbolic families.}\label{tab:appendix_inventory}\\
\toprule
Family & Atlas role & Temporal profile & Note \\
\midrule
\endfirsthead
\toprule
Family & Atlas role & Temporal profile & Note \\
\midrule
\endhead
\symentry{ساغر}{Saaghar}{Wine Cup} & Core / Wine-Tavern Ritual & Rising & A vessel image that remains central even as late lyric clustering intensifies. \\
\symentry{پیمانه}{Peymaaneh}{Wine Vessel} & Core / Wine-Tavern Ritual & Rising & A measure-and-intoxication family that strengthens with later lyric concentration. \\
\symentry{گلشن}{Golshan}{Garden} & Core / Floral-Vegetal & Rising & Garden-space imagery expands while remaining lexically transparent at the family level. \\
\symentry{خرقه}{Kherqe}{Sufi Robe} & Core / Ascetic-Mystical & Rising & A late-strengthening ritual family that becomes structurally prominent within the mystical field. \\
\symentry{خرابات}{Kharaabaat}{Tavern} & Core / Ascetic-Mystical & Peak Middle & A tavern family marking the field's recurrent traffic with wine imagery. \\
\symentry{بنفشه}{Banafsheh}{Violet} & Core / Body-Beloved & Declining & A beloved-image family weighted more strongly toward earlier and middle centuries. \\
\symentry{خاقان}{Khaaqaan}{Sovereign} & Referential / Courtly Bridge & Declining & A prestige head retained as a distinct courtly family rather than diffused into imagistic clusters. \\
\symentry{فرخ}{Farrokh}{Auspicious} & Core / Mixed Field & Declining & Auspicious prestige language that loses weight across the later chronology. \\
\symentry{شب}{Shab}{Night} & Core / Mixed Field & Stable & A durable lyric anchor with wide chronological spread. \\
\symentry{روز}{Ruz}{Day} & Core / Mixed Field & Stable & A companion temporal anchor that remains broadly distributed. \\
\symentry{خاک}{Khaak}{Earth} & Core / Mixed Field & Stable & Ethical and metaphysical weight gathers around a highly persistent elemental head. \\
\symentry{دریا}{Daryaa}{Sea} & Core / Water-Sea & Stable & A broad elemental family that remains recognizably tied to the same symbolic actor. \\
\symentry{آتش}{Aatash}{Fire} & Core / Light-Fire & Stable & An enduring intensity family linking flame, affect, and illumination. \\
\bottomrule
\end{longtable}
\par\endgroup

\subsection{Field structure and robustness}

Table~\ref{tab:appendix_fields} relocates the family illustrations removed from the main community graph into tabular form. The wine-tavern field is the most persistent large field, the floral-vegetal field shows the strongest internal splitting pressure, and the ascetic-mystical field is most often pulled toward wine imagery.

\begin{table}[!t]
\caption{Supplementary field inventory and structural tendencies.}
\label{tab:appendix_fields}
\centering
\footnotesize
\begin{tabularx}{\linewidth}{@{}p{0.22\linewidth}p{0.34\linewidth}p{0.12\linewidth}X@{}}
\toprule
Field & Illustrative families & Mean active-member ratio & Summary \\
\midrule
Wine-Tavern Ritual & \symgloss{Saaghar}{Wine Cup}; \symgloss{Baadeh}{Wine}; \symgloss{Qadah}{Goblet} & 0.990 & The most persistent large field in the atlas, with low split pressure and strong late continuity. \\
Floral-Vegetal & \symgloss{Gol}{Rose}; \symgloss{Golshan}{Garden}; \symgloss{Sarv}{Cypress} & 0.965 & The field remains active throughout the chronology but shows the strongest internal differentiation. \\
Body-Beloved & \symgloss{Banafsheh}{Violet}; \symgloss{Sonbol}{Hyacinth}; \symgloss{Meshkin}{Musky} & 0.960 & Bodily beauty imagery persists strongly while repeatedly leaning toward the floral field. \\
Ascetic-Mystical & \symgloss{Kherqe}{Sufi Robe}; \symgloss{Meykadeh}{Wine House}; \symgloss{Kharaabaat}{Tavern} & 0.857 & The most visibly intermediary field, with recurrent outward pull toward wine-tavern ritual. \\
Water-Sea & \symgloss{Daryaa}{Sea}; \symgloss{Bahr}{Sea}; \symgloss{Qolzum}{Sea} & 0.980 & A highly continuous elemental field with little internal fragmentation. \\
Light-Fire & \symgloss{Aatash}{Fire}; \symgloss{Forugh}{Glow}; \symgloss{Partow}{Radiance} & 0.985 & An unusually stable field whose members retain a common intensity profile across bins. \\
Mixed Field & \symgloss{Shab}{Night}; \symgloss{Ruz}{Day}; \symgloss{Khaak}{Earth} & 0.885 & A durable contact zone that repeatedly absorbs bridge attachment and historical retargeting. \\
\bottomrule
\end{tabularx}
\end{table}

Across loose, base, strict, and poet-balanced reconstructions, the direction of the central dynamic claims remains stable: core modularity rises, cross-scope linkage declines, and sacred-prophetic attachment strengthens. The most cautious result continues to be courtly weakening, which remains negative in the base and poet-balanced reconstructions but flattens under looser and stricter activity requirements. The smallest high-confidence-only subgraph preserves the broad tendency toward separation, but it becomes too sparse to sustain the full contrast structure on its own.

\subsection{Supplementary graph view}

Figure~A1 adds one more graph view without crowding the main text. It contrasts an earlier dense core snapshot with a later one. The later structure is more differentiated, but it remains continuous with the earlier network.

\begin{figure}[!t]
\centering
\includegraphics[width=0.90\linewidth]{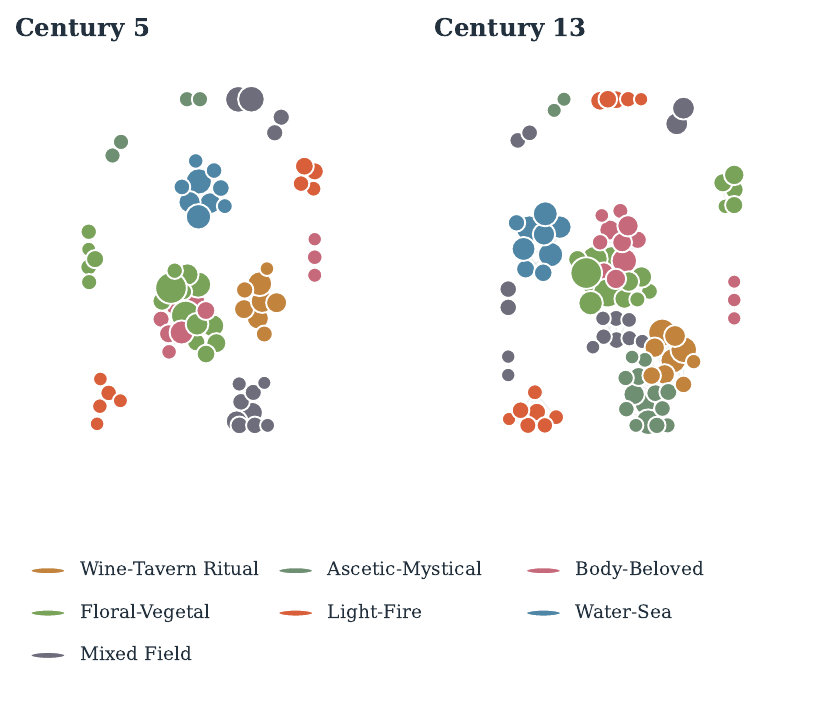}
\manualfigcaption{\textbf{Figure A1.}\hskip.7em Supplementary early and late core snapshots. Community colors follow Figure 3; the later snapshot shows the sharper differentiation of field structure discussed in the main text.}
\end{figure}

\subsection{Supplementary poet-level evidence}

The threshold-based core poet pool used in Figure~7 contains $36$ poets with at least $10{,}000$ verses and $1{,}000$ symbolic-family occurrences. This rule is coverage-driven rather than prestige-driven. It includes Saadi, Khaaqani, and Vahshi Bafqi, but excludes Hafez, whose mapped corpus in the present dataset reaches $5{,}183$ verses and falls below both the core and sensitivity floors. A broader $58$-poet sensitivity pool at $8{,}000$ verses and $500$ symbolic occurrences leaves the century-effect contrast unchanged.

Tables~\ref{tab:appendix_poet_profiles}--\ref{tab:appendix_poet_neighbors} provide the full verse-normalized field profiles, cluster assignments, and nearest symbolic neighbors for the core pool. Figure~A2 adds a low-dimensional view of the same standardized field space. The first two principal components capture $74.0\%$ of the variance and make the cluster geometry easier to inspect: Khaaqani and Vahshi Bafqi remain close to the symbolic center, whereas Saeb, Jahan Malek Khatun, and Ashofteh Shirazi separate most strongly from the pooled mean profile.

\begingroup
\footnotesize
\setlength{\tabcolsep}{2.4pt}
\begin{longtable}{@{}p{0.20\linewidth}crrrrrrrrrr@{}}
\caption{Verse-normalized poet-field profiles for the threshold-based core poet pool. The pool retains poets with at least 10{,}000 verses and 1{,}000 symbolic-family occurrences. WT = Wine-Tavern, AM = Ascetic-Mystical, BB = Body-Beloved, FV = Floral-Vegetal, LF = Light-Fire, WS = Water-Sea, MF = Mixed Field. Cluster assignments are reported separately in Table~\ref{tab:appendix_poet_neighbors}.}\label{tab:appendix_poet_profiles}\\
\toprule
Poet & C. & Verses & Sym. & WT & AM & BB & FV & LF & WS & MF \\
\midrule
\endfirsthead
\toprule
Poet & C. & Verses & Sym. & WT & AM & BB & FV & LF & WS & MF \\
\midrule
\endhead
Amir Khosrow Dehlavi & 7 & 21,599 & 19,843 & 1714.9 & 170.8 & 1387.6 & 885.7 & 333.3 & 153.7 & 4541.0 \\
Jahan Malek Khatun & 8 & 12,276 & 12,720 & 165.4 & 31.8 & 1376.7 & 1770.9 & 432.6 & 127.1 & 6457.3 \\
Khwaju Kermani & 8 & 29,301 & 19,768 & 1128.3 & 158.4 & 1079.5 & 771.0 & 414.3 & 297.9 & 2897.2 \\
Salman Savaji & 8 & 13,129 & 10,573 & 844.7 & 207.9 & 1352.7 & 1132.6 & 401.4 & 320.7 & 3793.1 \\
Seyf Farghani & 8 & 10,279 & 8,205 & 803.6 & 74.9 & 1192.7 & 917.4 & 583.7 & 180.0 & 4230.0 \\
Jami & 9 & 39,753 & 26,396 & 1129.7 & 218.3 & 999.2 & 739.3 & 441.0 & 203.8 & 2908.7 \\
Ahli Shirazi & 10 & 16,064 & 13,948 & 1012.8 & 262.7 & 1194.6 & 1289.8 & 692.2 & 180.5 & 4050.0 \\
Saeb & 11 & 77,570 & 89,285 & 3250.2 & 213.0 & 1624.6 & 1338.4 & 666.5 & 771.0 & 3646.5 \\
Hazin Lahiji & 12 & 14,480 & 13,904 & 2138.8 & 367.4 & 1195.4 & 953.0 & 627.1 & 341.9 & 3978.6 \\
Ashofteh Shirazi & 13 & 12,767 & 11,349 & 1178.8 & 604.7 & 1294.7 & 1077.8 & 623.5 & 246.7 & 3863.1 \\
Amir Mu'izzi & 5 & 18,406 & 9,220 & 305.3 & 31.0 & 690.5 & 420.0 & 439.5 & 286.3 & 2836.6 \\
Mas'ud Sa'd Salman & 5 & 15,048 & 8,109 & 536.9 & 20.6 & 534.3 & 579.5 & 585.5 & 249.2 & 2882.8 \\
Sana'i & 5 & 26,079 & 11,615 & 265.3 & 126.2 & 524.9 & 300.6 & 380.4 & 148.8 & 2707.5 \\
Anvari & 6 & 13,308 & 5,946 & 248.7 & 32.3 & 401.3 & 310.3 & 259.2 & 214.9 & 3001.2 \\
Attar & 6 & 94,934 & 38,521 & 257.3 & 81.3 & 316.6 & 218.8 & 507.4 & 335.7 & 2340.5 \\
Khaaqani & 6 & 17,434 & 9,568 & 448.5 & 122.2 & 792.1 & 311.5 & 513.9 & 273.6 & 3026.3 \\
Hakim Nezari & 7 & 18,402 & 10,549 & 1431.4 & 172.8 & 501.6 & 380.9 & 301.6 & 144.0 & 2800.2 \\
Mowlana & 7 & 66,196 & 34,220 & 638.3 & 117.8 & 618.9 & 400.0 & 534.2 & 346.1 & 2514.2 \\
Saadi & 7 & 16,159 & 6,939 & 140.5 & 104.0 & 648.6 & 547.1 & 199.9 & 125.0 & 2529.2 \\
Ibn Yamin & 8 & 13,792 & 7,811 & 525.7 & 87.7 & 855.6 & 710.6 & 470.6 & 249.4 & 2763.9 \\
Owhadi & 8 & 14,683 & 8,760 & 345.3 & 149.2 & 869.0 & 506.0 & 397.7 & 96.0 & 3602.8 \\
Mohtasham Kashani & 10 & 10,716 & 5,985 & 286.5 & 42.9 & 911.7 & 691.5 & 465.7 & 328.5 & 2858.3 \\
Vahshi Bafqi & 10 & 10,596 & 6,058 & 380.3 & 100.0 & 699.3 & 668.2 & 448.3 & 260.5 & 3160.6 \\
Bidel Dehlavi & 11 & 32,333 & 17,434 & 462.1 & 63.7 & 834.1 & 634.6 & 458.0 & 586.1 & 2353.3 \\
Fayyaz Lahiji & 11 & 10,204 & 7,630 & 612.5 & 128.4 & 1004.5 & 1126.0 & 691.9 & 472.4 & 3441.8 \\
Feyz Kashani & 11 & 11,240 & 8,111 & 815.8 & 191.3 & 781.1 & 409.3 & 556.0 & 232.2 & 4230.4 \\
Molla Ahmad Naraqi & 13 & 11,356 & 5,202 & 548.6 & 115.4 & 443.8 & 383.9 & 685.1 & 250.1 & 2153.9 \\
Qa'ani & 13 & 20,343 & 9,561 & 532.9 & 72.3 & 884.3 & 516.1 & 390.3 & 336.2 & 1967.8 \\
Reza-Qoli Khan Hedayat & 13 & 10,028 & 5,179 & 444.8 & 324.1 & 386.9 & 240.3 & 523.5 & 431.8 & 2813.1 \\
Ferdowsi & 4 & 49,610 & 12,502 & 146.1 & 2.2 & 192.7 & 136.3 & 111.3 & 127.4 & 1804.1 \\
Iranshahan & 5 & 20,825 & 5,573 & 121.0 & 1.9 & 155.6 & 123.9 & 126.3 & 253.5 & 1893.9 \\
Nasir Khusraw & 5 & 10,944 & 3,039 & 145.3 & 50.3 & 325.3 & 287.8 & 246.7 & 120.6 & 1600.9 \\
Nizami & 6 & 27,768 & 9,392 & 266.9 & 32.1 & 415.9 & 361.6 & 315.8 & 189.4 & 1800.6 \\
Elhami Kermanshahi & 13 & 16,795 & 5,371 & 216.7 & 36.9 & 350.7 & 206.0 & 235.8 & 111.9 & 2039.9 \\
Safi Alishah & 13 & 44,939 & 10,124 & 677.4 & 72.1 & 182.7 & 53.6 & 239.4 & 163.3 & 864.3 \\
Adib al-Mamalek & 14 & 15,562 & 5,894 & 468.4 & 68.8 & 455.6 & 447.9 & 286.6 & 241.0 & 1819.2 \\
\bottomrule
\end{longtable}
\par\endgroup

\begingroup
\footnotesize
\setlength{\tabcolsep}{3.0pt}
\begin{longtable}{@{}p{0.20\linewidth}p{0.18\linewidth}p{0.22\linewidth}p{0.22\linewidth}r@{}}
\caption{Cluster membership and nearest symbolic neighbors in standardized field space. Smaller mean-distance ranks indicate more central positions, while larger ranks indicate more outlying profiles.}\label{tab:appendix_poet_neighbors}\\
\toprule
Poet & Cluster & Nearest neighbor & Second neighbor & Mean-dist. rank \\
\midrule
\endfirsthead
\toprule
Poet & Cluster & Nearest neighbor & Second neighbor & Mean-dist. rank \\
\midrule
\endhead
Amir Khosrow Dehlavi & Lyric-amplifying & Salman Savaji & Jami & 31 \\
Jahan Malek Khatun & Lyric-amplifying & Seyf Farghani & Salman Savaji & 35 \\
Khwaju Kermani & Lyric-amplifying & Jami & Ibn Yamin & 9 \\
Salman Savaji & Lyric-amplifying & Khwaju Kermani & Jami & 24 \\
Seyf Farghani & Lyric-amplifying & Feyz Kashani & Ibn Yamin & 22 \\
Jami & Lyric-amplifying & Khwaju Kermani & Ibn Yamin & 12 \\
Ahli Shirazi & Lyric-amplifying & Seyf Farghani & Salman Savaji & 32 \\
Saeb & Lyric-amplifying & Hazin Lahiji & Fayyaz Lahiji & 36 \\
Hazin Lahiji & Lyric-amplifying & Ahli Shirazi & Ashofteh Shirazi & 33 \\
Ashofteh Shirazi & Lyric-amplifying & Hazin Lahiji & Ahli Shirazi & 34 \\
Amir Mu'izzi & Mediating-balanced & Vahshi Bafqi & Mohtasham Kashani & 4 \\
Mas'ud Sa'd Salman & Mediating-balanced & Amir Mu'izzi & Vahshi Bafqi & 10 \\
Sana'i & Mediating-balanced & Anvari & Nizami & 7 \\
Anvari & Mediating-balanced & Nizami & Elhami Kermanshahi & 15 \\
Attar & Mediating-balanced & Mowlana & Amir Mu'izzi & 16 \\
Khaaqani & Mediating-balanced & Mowlana & Amir Mu'izzi & 3 \\
Hakim Nezari & Mediating-balanced & Sana'i & Jami & 19 \\
Mowlana & Mediating-balanced & Khaaqani & Attar & 5 \\
Saadi & Mediating-balanced & Sana'i & Anvari & 18 \\
Ibn Yamin & Mediating-balanced & Vahshi Bafqi & Mohtasham Kashani & 2 \\
Owhadi & Mediating-balanced & Sana'i & Vahshi Bafqi & 13 \\
Mohtasham Kashani & Mediating-balanced & Ibn Yamin & Vahshi Bafqi & 6 \\
Vahshi Bafqi & Mediating-balanced & Ibn Yamin & Amir Mu'izzi & 1 \\
Bidel Dehlavi & Mediating-balanced & Qa'ani & Mohtasham Kashani & 25 \\
Fayyaz Lahiji & Mediating-balanced & Mohtasham Kashani & Salman Savaji & 27 \\
Feyz Kashani & Mediating-balanced & Khaaqani & Vahshi Bafqi & 17 \\
Molla Ahmad Naraqi & Mediating-balanced & Mowlana & Mas'ud Sa'd Salman & 20 \\
Qa'ani & Mediating-balanced & Mohtasham Kashani & Amir Mu'izzi & 8 \\
Reza-Qoli Khan Hedayat & Mediating-balanced & Mowlana & Attar & 26 \\
Ferdowsi & Sparse-anchor & Iranshahan & Elhami Kermanshahi & 30 \\
Iranshahan & Sparse-anchor & Ferdowsi & Elhami Kermanshahi & 28 \\
Nasir Khusraw & Sparse-anchor & Elhami Kermanshahi & Nizami & 23 \\
Nizami & Sparse-anchor & Adib al-Mamalek & Nasir Khusraw & 14 \\
Elhami Kermanshahi & Sparse-anchor & Nasir Khusraw & Nizami & 21 \\
Safi Alishah & Sparse-anchor & Nasir Khusraw & Elhami Kermanshahi & 29 \\
Adib al-Mamalek & Sparse-anchor & Nizami & Nasir Khusraw & 11 \\
\bottomrule
\end{longtable}
\par\endgroup

\begin{figure}[!t]
\centering
\includegraphics[width=0.90\linewidth]{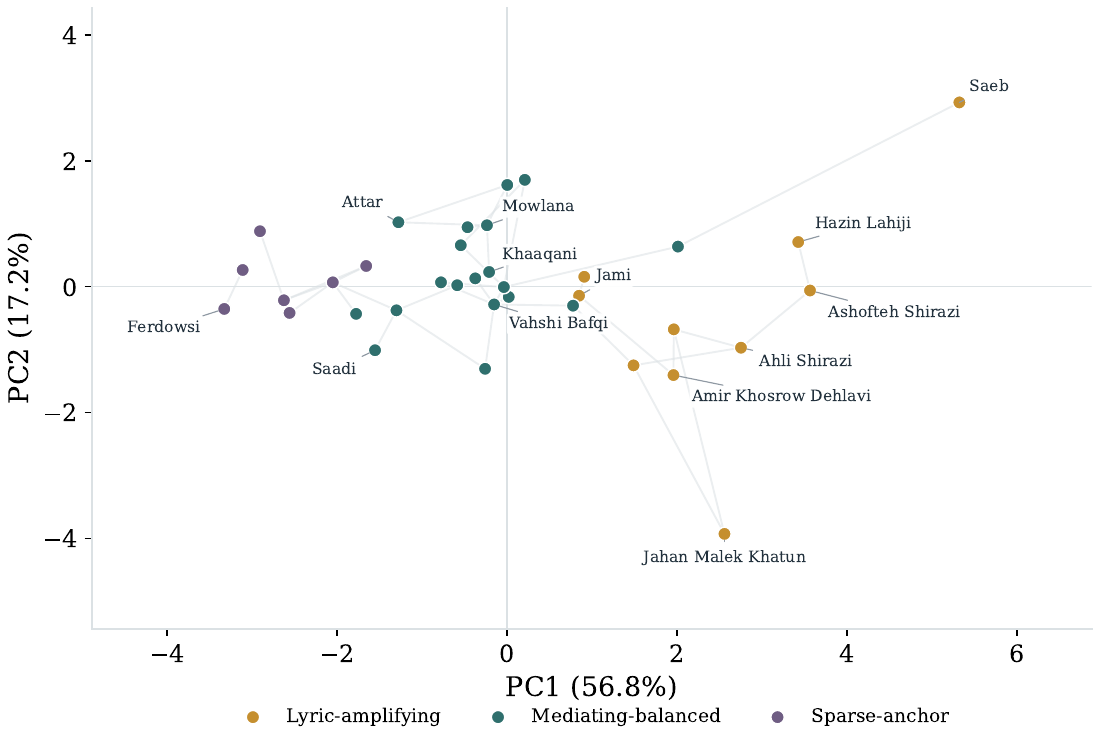}
\manualfigcaption{\textbf{Figure A2.}\hskip.7em Supplementary projection of poet similarity space. Points show the threshold-based core poet pool in the first two principal components of the standardized field profiles; faint lines connect nearest neighbors, and colors follow the three clusters shown in Figure~7.}
\end{figure}

\end{document}